# The Numerical Control Design for a Pair of Dubin's Vehicles

Heru Tjahjana[1], Iwan Pranoto[2], Hari Muhammad[3], J. Naiborhu[4], and Miswanto[5]

[1,2,4,5]Mathematics
Institut Teknologi Bandung, Indonesia
[1]Permanent address: Mathematics
Diponegoro University, Indonesia

e-mail:[1] heru_tjahjana@students.itb.ac.id
[2] pranoto@math.itb.ac.id
[4] janson@math.itb.ac.id
[5] miswanto@students.itb.ac.id

[3]Aeronautics and Astronautics
Institut Teknologi Bandung, Bandung, Indonesia
e-mail: harmad@ae.itb.ac.id

**Abstract**

In this paper, a model of a pair of Dubin's vehicles is considered. The vehicles move from an initial position and orientation to final position and orientation. A long the motion, the two vehicles are not allowed to collide however the two vehicles cann't to far each other. The optimal control of the vehicle is found using the Pontryagin's Maximum Principle (PMP). This PMP leads to a Hamiltonian system consisting of a system of differential equation and its adjoint. The originally differential equation has initial and final condition but the adjoint system doesn't have one. The classical difficulty is solved numerically by the greatest gradient descent method. Some simulation results are presented in this paper.

## 1 Introduction

The swarm behavior in nature is interesting by itself, but in this current paper modeling the multi-vehicle systems together with their designed optimal control is considered. Multi-agent system and swarm phenomena have been studied extensively. Many studies regarding these two topics have been done and published, but all of them focus on the local swarm behavior or something very far from transportation situations[6]-[9]. They cannot be applied to the problems such as aircraft or ship convoy. In this paper, the optimal control of a pair of Dubin's vehicles is designed. The previous research about swarm modeling through optimal control can be found in [1]-[5]. The previous paper which expos Dubin's vehicles can be found in [10]-[12].

## 2 A Pair of Dubin's Vehicles Model

Consider a pair of Dubin's vehicles model. The model of first vehicles is given as follows

$$\dot{x}_1 = (\sin x_3)u_1$$
$$\dot{x}_2 = (\cos x_3)u_1 \qquad (1)$$
$$\dot{x}_3 = u_2$$

And the second vehicle is given as

$$\dot{y}_1 = (\sin y_3)v_1$$
$$\dot{y}_2 = (\cos y_3)v_1 \qquad (2)$$
$$\dot{y}_3 = v_2$$

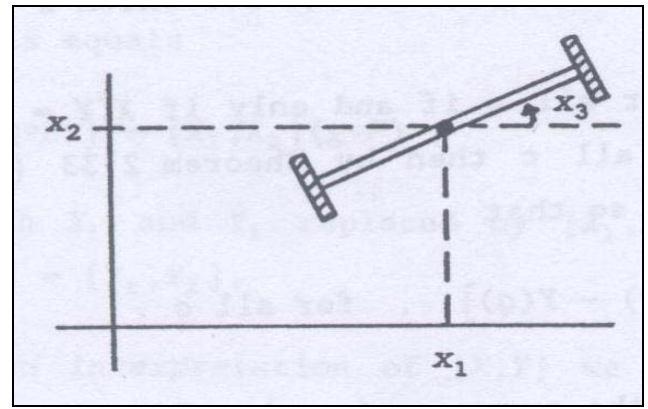

**Figure 1: Model of Dubin's Vehicle**

The functional cost that must be minimized is

$$J = \frac{1}{2}\int_0^T \delta u_1^2 + \delta u_2^2 + \delta v_1^2 + \delta v_2^2$$
$$+ \beta(x_1^2 + x_2^2) + \alpha(y_1^2 + y_2^2) \qquad (3)$$
$$+ \frac{\rho}{(x_1 - y_1)^2 + (x_2 - y_2)^2} dt$$

The Hamiltonian function is

$$H = p_1(u_1 \sin x_3) + p_2(u_1 \cos x_3) + p_3 u_2 +$$
$$+ p_4(v_1 \sin y_3) + p_5(v_1 \cos y_3) + p_6 v_2$$
$$+ \frac{1}{2} p_0 \delta u_1^2 + \frac{1}{2} p_0 \delta u_2^2 + \frac{1}{2} p_0 \delta v_1^2$$
$$+ \frac{1}{2} p_0 \delta v_2^2 + \frac{1}{2} p_0 \beta x_1^2 + \frac{1}{2} p_0 \beta x_2^2 \quad (4)$$
$$+ \frac{1}{2} p_0 \alpha y_1^2 + \frac{1}{2} p_0 \alpha y_2^2$$
$$+ \frac{1}{2} \frac{\rho p_0}{(x_1 - y_1)^2 + (x_2 - y_2)^2}$$

The Hamiltonian system can be expressed as

$$\frac{\partial H}{\partial p_1} = \dot{x}_1 = (\sin x_3) u_1$$

$$\frac{\partial H}{\partial p_2} = \dot{x}_2 = (\cos x_3) u_1$$

$$\frac{\partial H}{\partial p_3} = \dot{x}_3 = u_2$$

$$\frac{\partial H}{\partial x_1} = -\dot{p}_1 = p_0 \beta x_1 - \frac{1}{2} \frac{\rho p_0 (2x_1 - 2y_1)}{\left((x_1 - y_1)^2 + (x_2 - y_2)^2\right)}$$

$$\frac{\partial H}{\partial x_2} = -\dot{p}_2 = p_0 \beta x_2 - \frac{1}{2} \frac{\rho p_0 (2x_2 - 2y_2)}{\left((x_1 - y_1)^2 + (x_2 - y_2)^2\right)}$$

$$\frac{\partial H}{\partial x_3} = -\dot{p}_3 = p_1 u_1 \cos x_3 - p_2 u_1 \sin x_3$$

$$\frac{\partial H}{\partial u_1} = 0 = p_1 \sin x_3 + p_2 \cos x_3 + p_0 \delta u_1$$

$$\frac{\partial H}{\partial u_2} = 0 = P_3 + p_0 \delta u_2$$

$$\frac{\partial H}{\partial p_4} = \dot{y}_1 = (\sin y_3) v_1$$

$$\frac{\partial H}{\partial p_5} = \dot{y}_2 = (\cos y_3) v_1$$

$$\frac{\partial H}{\partial p_6} = \dot{y}_3 = v_2$$

$$\frac{\partial H}{\partial y_1} = -\dot{p}_4 = p_0 \alpha y_1 - \frac{1}{2} \frac{\rho p_0 (-2x_1 + 2y_1)}{\left((x_1 - y_1)^2 + (x_2 - y_2)^2\right)}$$

$$\frac{\partial H}{\partial y_2} = -\dot{p}_5 = p_0 \alpha y_2 - \frac{1}{2} \frac{\rho p_0 (-2x_2 + 2y_2)}{\left((x_1 - y_1)^2 + (x_2 - y_2)^2\right)} \quad (5)$$

$$\frac{\partial H}{\partial y_3} = -\dot{p}_6 = p_4 v_1 \cos y_3 - p_5 v_1 \sin y_3$$

$$\frac{\partial H}{\partial v_1} = 0 = p_4 \sin y_3 + p_5 \cos y_3 + p_0 \delta u_1$$

$$\frac{\partial H}{\partial u_2} = 0 = P_6 + p_0 \delta v_2$$

Consider the Hamiltonian system (5), the adjoint variables or co-state variables $p_i$ appear in the system. The original state variables are companied by the co-state variables. Here, the problem is the co-state variable don't have initial value or initial condition. This fundamental difficulty can be solved by greatest gradient descent method or sometime is called as shooting method. Through Pontryagin Maximum Principle, the optimal control of two vehicles can be found. If the optimal control of two vehicles substitute to the Hamiltonian system, then the differential equation system (5) equivalent with

$$\dot{x}_1 = (\sin x_3)(p_1 \sin^2 x_3 + p_2 \sin x_3 \cos x_3)\left(\frac{1}{\delta}\right)$$

$$\dot{x}_2 = (\cos x_3)(p_1 \sin^2 x_3 + p_2 \sin x_3 \cos x_3)\left(\frac{1}{\delta}\right)$$

$$\dot{x}_3 = \frac{P_3}{\delta}$$

$$-\dot{p}_1 = -\beta x_1 + \frac{1}{2} \frac{\rho(2x_1 - 2y_1)}{\left((x_1 - y_1)^2 + (x_2 - y_2)^2\right)}$$

$$-\dot{p}_2 = -\beta x_2 + \frac{1}{2} \frac{\rho(2x_2 - 2y_2)}{\left((x_1 - y_1)^2 + (x_2 - y_2)^2\right)}$$

$$-\dot{p}_3 = p_1(\cos x_3)(p_1 \sin x_3 + p_2 \cos x_3)\left(\frac{1}{\delta}\right) -$$
$$p_2(\sin x_3)(p_1 \sin x_3 + p_2 \cos x_3)\left(\frac{1}{\delta}\right)$$

$$\dot{y}_1 = (\sin x_3)(p_4 \sin^2 x_3 + p_5 \sin y_3 \cos y_3)\left(\frac{1}{\delta}\right)$$

$$\dot{y}_2 = (\cos x_3)(p_1 \sin^2 x_3 + p_2 \sin x_3 \cos x_3)\left(\frac{1}{\delta}\right)$$

$$\dot{y}_3 = \frac{P_6}{\delta}$$

$$-\dot{p}_4 = -\alpha y_1 + \frac{1}{2} \frac{\rho(-2x_1 + 2y_1)}{\left((x_1 - y_1)^2 + (x_2 - y_2)^2\right)}$$

$$-\dot{p}_5 = -\alpha y_2 + \frac{1}{2} \frac{\rho(-2x_2 + 2y_2)}{\left((x_1 - y_1)^2 + (x_2 - y_2)^2\right)} \quad (6)$$

$$-\dot{p}_6 = p_4(\cos y_3)(p_4 \sin y_3 + p_5 \cos y_3)\left(\frac{1}{\delta}\right) -$$
$$p_5(\sin x_3)(p_4 \sin y_3 + p_5 \cos y_3)\left(\frac{1}{\delta}\right)$$

If the initial condition for vehicle 1 and vehicle 2 are (5.5,0,0) and (15.5,0,0) respectively and the terminal condition are given (0,0.2,0) and (9.8,0,0) respectively, then the simulation result as follows

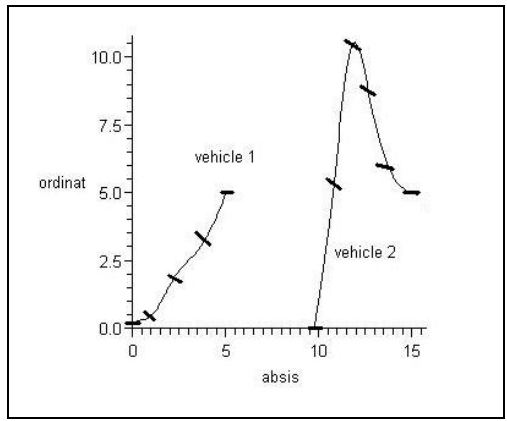

**Figure 2: Optimal Trajectory of Dubin's Vehicles**

### 3 Conclusions

The greatest gradient descent method can be used to solve the problem that the originally differential equation has initial and final condition but the adjoint system doesn't have one. Through this method, the optimal control and optimal trajectory of a pair of Dubin's vehicles can be found.